\documentclass[sigconf]{acmart}

\usepackage{subfigure}

\PassOptionsToPackage{hyphens}{url}
\usepackage{hyperref}
\usepackage[nobiblatex]{xurl}
\usepackage{url}

\usepackage{times}
\usepackage{latexsym}

\usepackage{graphicx}
\usepackage{caption}
\usepackage{CJKutf8}
\usepackage{microtype}
\usepackage{amsfonts}
\usepackage{amsmath}
\usepackage{enumitem}
\usepackage{multirow,multicol,makecell,array}
\usepackage[linesnumbered,commentsnumbered,ruled]{algorithm2e}

\AtBeginDocument{%
  \providecommand\BibTeX{{%
    \normalfont B\kern-0.5em{\scshape i\kern-0.25em b}\kern-0.8em\TeX}}}

\setcopyright{iw3c2w3}

\copyrightyear{2021}
\acmYear{2021} 
\acmConference[WWW '21]{Proceedings of the Web Conference 2021}{April 19--23, 2021}{Ljubljana, Slovenia} 
\acmBooktitle{Proceedings of the Web Conference 2021 (WWW '21), April 19--23, 2021, Ljubljana, Slovenia}
\acmPrice{}
\acmDOI{10.1145/3442381.3449838}
\acmISBN{978-1-4503-8312-7/21/04}

\begin{document}

\title{Controllable and Diverse Text Generation in E-commerce}


\author{Huajie Shao}
\authornote{Part of work was completed at Alibaba Group.}
\affiliation{%
  \institution{1. University of Illinois at Urbana Champaign}
  \country{2. Alibaba Group}
}
\email{hshao5@illinois.edu}

\author{Jun Wang}
\affiliation{%
  \country{Alibaba Group}
 }
 \email{jun.w@alibaba-inc.com}

\author{Haohong Lin}
\affiliation{%
  \country{Zhejiang University}
}
\email{lhh2017@zju.edu.cn}

\author{Xuezhou Zhang}
\affiliation{%
  \country{University of Wisconsin, Madison}
}
\email{zhangxz1123@cs.wisc.edu}

\author{Aston Zhang}
\affiliation{%
  \country{University of Illinois at Urbana Champaign}
}
\email{lzhang74@illinois.edu}

\author{Heng Ji}
\affiliation{%
  \country{University of Illinois at Urbana Champaign}
}
 \email{hengji@illinois.edu}

\author{Tarek Abdelzaher}
\affiliation{%
  \country{University of Illinois at Urbana Champaign}
}
\email{zaher@illinois.edu}

%
%
%
%
%

\renewcommand{\shortauthors}{Huajie Shao and Jun Wang, et al.}

%
\begin{abstract}
In E-commerce, a key challenge in text generation is to find a good trade-off between word diversity and accuracy (relevance) in order to make generated text appear more natural and human-like. In order to improve the relevance of generated results, conditional text generators were developed that use input keywords or attributes to produce the corresponding text. Prior work, however, do not finely control the diversity of automatically generated sentences. For example, it does not control the order of keywords to put more relevant ones first. Moreover, it does not explicitly control the balance between diversity and accuracy. To remedy these problems, we propose a fine-grained controllable generative model, called~\textit{Apex}, that uses an algorithm borrowed from automatic control (namely, a variant of the \textit{proportional, integral, and derivative (PID) controller}) to precisely manipulate the diversity/accuracy trade-off of generated text. The algorithm is injected into a Conditional Variational Autoencoder (CVAE), allowing \textit{Apex} to control both (i) the order of keywords in the generated sentences (conditioned on the input keywords and their order), and (ii) the trade-off between diversity and accuracy. Evaluation results on real world datasets~\footnote{The sampled data is publicly available at \url{https://github.com/paper-data-open/Text-Gen-Ecommerce}} show that the proposed method outperforms existing generative models in terms of diversity and relevance. Moreover, it achieves about 97\% accuracy in the control of the order of keywords. 

\textit{Apex} is currently deployed to generate production descriptions and item recommendation reasons in Taobao\footnote{https://www.taobao.com/}, the largest E-commerce platform in China. The A/B production test results show that our method improves click-through rate (CTR) by 13.17\% compared to the existing method for production descriptions. For item recommendation reason, it is able to increase CTR by 6.89\% and 1.42\% compared to user reviews and top-K item recommendation without reviews, respectively.
\end{abstract}

\maketitle

{
\section{Introduction}
\label{sec:intro}
Text generation has been widely used in a variety of natural language processing (NLP) applications, such as review writing assistance~\cite{zang2017towards,wang2020reviewrobot}, production description generation~\cite{wang2017statistical}, and dialogue generation~\cite{li2017adversarial}. In E-commerce, an attractive and accurate product description is crucial to convince customers to click or purchase the recommended items.  Thus, how to strike the right balance between diversity and accuracy (relevance)\footnote{relevance and accuracy are interchangeable in this paper} of generated text~\cite{zhang2018generating} remains a key challenge in text generation. As we know, diversity can help the generated text seem more natural and human-like, while accuracy means the generated text is relevant to the target. In general, techniques that improve diversity reduce accuracy, whereas techniques that improve accuracy often produce the same expressions repeatedly. The contribution of this paper is to develop a controllable generative model that manipulates both the diversity and accuracy of generated sentences to attain both goals at the same time.

Recently, deep generative models have been proposed to diversify generated text on the basis of Variational Autoencoders (VAEs)~\cite{hu2017toward,wang2019topic} and generative adversarial networks (GAN)~\cite{yu2017seqgan,guo2018long,xu2018diversity,zhang2016generating}. To better control the relevance of generated results, researchers developed conditional text generators, conditioned on some keywords or attributes of items, user identities, or semantics~\cite{peng2018towards,hu2017toward,zang2017towards}. 

Past methods, however, do not manipulate the order of keywords nor accurately diversify the generated text. The order of keywords in the generated sentences plays an important role in attracting customers in E-commerce applications. It can provide most important production information to customers, and promote their interests to improve the likelihood of clicking or purchase. For example, when one recommends a skirt to a customer, one can write: ``The skirt is very \textbf{affordable}, yet \textbf{charming}'' if the customer prefers price to fashion. Alternatively, one can rephrase: ``The skirt is very  \textbf{charming}, yet \textbf{affortable}'' if the customer prefers fashion. This observation motivates us to control the order of keywords to generate various product descriptions in order to attract different customers.

Diversity makes text appear more human like. A sentence that conveys the intended meaning might be: ``The trousers are \textbf{comfortable} with \textbf{comfortable} fabric, and \textbf{comfortable} fit''. A more diverse description might say: ``The trousers are \textbf{comfortable} with \textbf{soft} fabric, and \textbf{relaxed} fit''. The key is to diversify while maintaining accuracy.

To this end, we propose a novel controllable text generation framework, called \textit{Apex}. \textit{Apex} combines Conditional Variational Autoencoders (CVAEs), illustrated in Fig.~\ref{fig:apex}, with a linear PI controller (as shown in Fig.~\ref{fig:pid}), a variant of proportional, integral, differential (PID) control algorithm~\cite{aastrom1993automatic}, to control the diversity and accuracy of generated text. Specifically, the CVAEs condition the keywords and their order appearing in the reference text on the conditional encoder with Transformer~\cite{vaswani2017attention} to control the diversity of generated sentences. However, CVAEs often suffer from the KL-vanishing problem for sequence generation models~\cite{shen2018improving,zhao2017learning,liu2019cyclical}. Namely, the KL-divergence term becomes zero and the decoder completely ignores latent features conditioned on input data. This is because the powerful decoder can directly learn information from other paths, such as the input keywords of the conditional encoder (red arrow) in Fig.~\ref{fig:apex}. On the other hand, KL divergence in CVAEs can affect the diversity and accuracy of generated text. A large KL-divergence can diversify generated text, but lower its accuracy. Hence, controlling the value of KL-divergence is of great importance to text generation.

Prior work, such as cost annealing and cyclical annealing~\cite{liu2019cyclical,bowman2015generating}, only focus on mitigating the KL-vanishing problem, but cannot explicitly control the value of KL-divergence. Besides, these methods cannot fully avert KL-vanishing because they blindly tune the weight on KL term without using feedback (from output KL-divergence) during model training. In practice, the feedback from KL-divergence can tell us when the KL-vanishing problem occurs so that we can change the weight on the KL term accordingly in the VAE objective function. Specifically, when KL-divergence becomes too small (thus hurting diversity), we need to reduce the weight of the KL term in the objective being optimized, allowing the optimization to favor larger KL-divergence values. Conversely, when KL-divergence becomes too large (thus possibly hurting accuracy), we increase its weight in the objective function, causing the optimization to favor smaller values. This is not unlike control of composition of chemical ingredients to optimize specific product qualities.

Inspired by control literature, we design a PI controller, a variant of PID control algorithm~\cite{aastrom1993automatic,aastrom2006advanced} as introduced in Section~\ref{sec:pid_intro}, that is novel in using KL-divergence feedback. The basic idea of PID controller is to calculate the error between a desired value (in this case, the desired KL-divergence) and the current actual value, then apply a correction in a direction that reduces that error. By doing this, it stabilizes the actual KL-divergence around a specified value, called {\em set point\/}. In this paper, the designed PI controller dynamically tunes the weight on KL term in the CVAE objective based on the output KL-divergence to stabilize it to a desired value. In this way, we can not only totally solve KL-vanishing problem, but also achieve an {\em adjustable trade-off\/} between diversity and accuracy of generated text via controlling KL-divergence.


We evaluate the performance of \textit{Apex} on the real-world dataset, collected from Taobao, a Chinese E-commerce platform. The experimental results illustrate that \textit{Apex} outperforms the state-of-the-art text generators in terms of diversity and accuracy. It can also control the order of keywords in the generated sentences with about 97\% accuracy. 

Importantly, \textit{Apex} has been deployed in Taobao E-commerce platform to generate production descriptions and item recommendation reasons. According to A/B tests, our method achieves an improvement of 13.17\% over the existing method in terms of CTR for production description generation. For item recommendation reason, it improves CTR by 6.89\% and 1.42\% compared to user reviews and top-K item recommendation without reviews, respectively.

\begin{figure}[!t]
\centering
\includegraphics[width= 3 in]{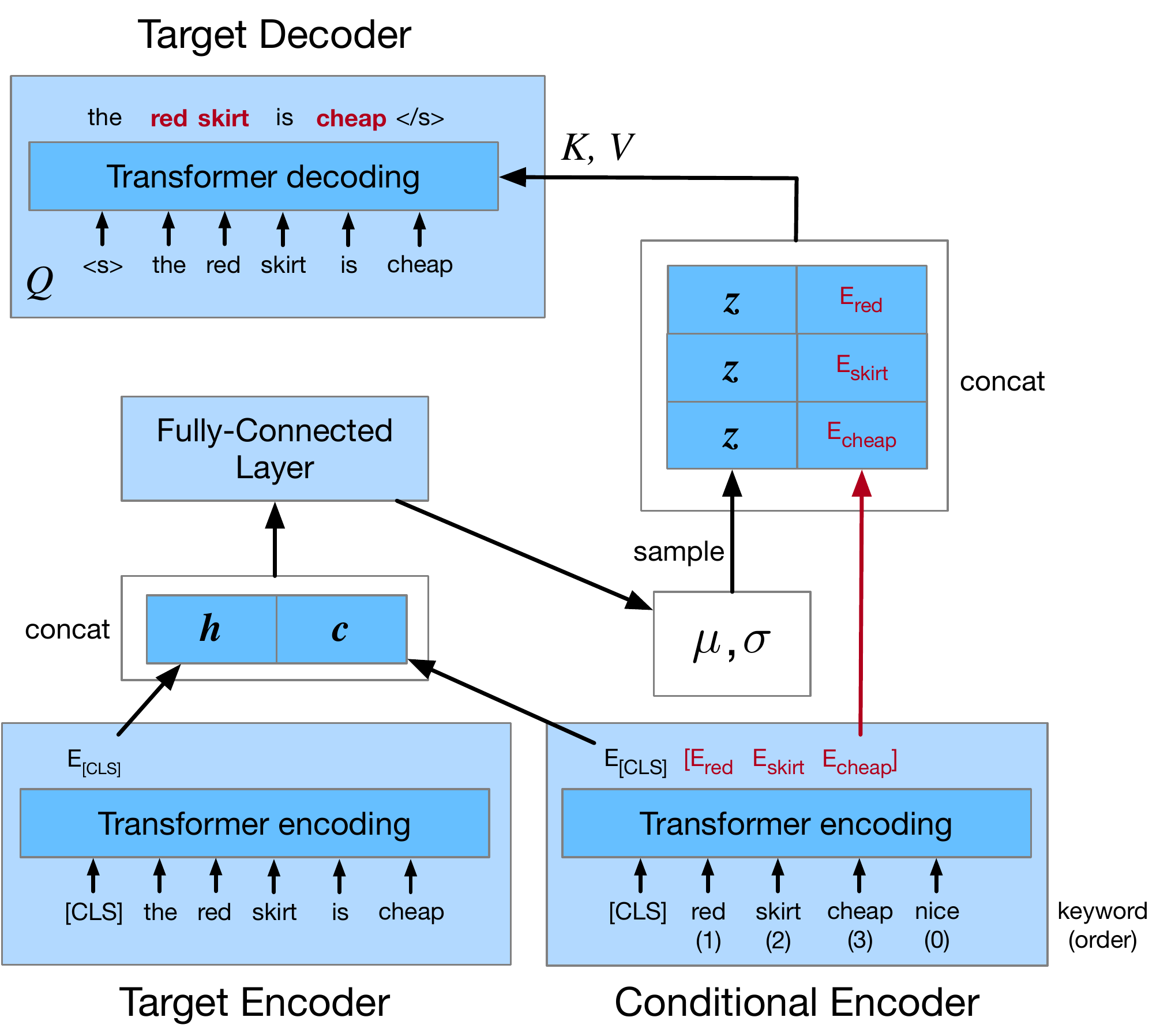}
\caption{The CVAEs component of the proposed \textit{Apex}.}
\vspace{-0.1 in}
\label{fig:apex}
\end{figure}


\section{Preliminaries}
\label{sec:problem}
We first introduce our text generation problem, and then present the background of two generative models, variational autoencoders (VAEs) and conditional variational autoencoders (CVAEs). Finally, we review the general PID control algorithm in automatic control theory.

\subsection{Problem}
The goal of this work is to generate text descriptions of items given selected keywords and their order. For a given item, let the set $Y_a= \{y_1,y_2, \dots, y_m\}$ denote the set of selected keywords, and $Y_p = \{p_1,p_2,\dots,p_m\}$ denote their corresponding order occurring in the reference text, such that $p_i$, $i \in \{1,2,\dots,m\}$, is the order of keyword $y_i$. Given $Y_a$ and $Y_p$ for the item, the goal is to automatically generate an accurate text description of the item, where the specified keywords are used (in meaningful descriptive sentences) and appear in the specified order.

To meet the above goal, we need to train a text generation model. To train a model, for each set of input keywords $Y_a$ of an item in the training data, we use a corresponding reference text with $n$ words, denoted by $X_{1:n} = \{x_1,x_2, \dots, x_n\}$. The keywords appear in  the order, $Y_p$, in the reference text. Overloading the order notation, let $p_i=0$ mean that the $i$-th keyword is \textit{not} in the reference text. As shown in Fig.~\ref{fig:apex}, the keyword, ``red'', is the first one among other keywords in the reference sentence. It is thus numbered as ``1'' in numerical order. For keyword ``nice'', it is numbered as ``0'' because it is not in the reference text. Note that, the reference text is used to train the CVAEs model to learn the conditional distribution of latent variable, denoted by $(\mathbf{\mu}, \mathbf{\sigma})$.

During testing, we use sampled data from the distribution of the latent variable, $(\mathbf{\mu}, \mathbf{\sigma})$, together with the input keywords $Y_a$ and their order $Y_p$ to generate text.

\subsection{Background of VAEs and CVAEs}
\label{subsec:vae} 
As one of the most popular deep generative models, VAEs~\citep{kingma2013auto,liu2017unsupervised} have been widely used in various applications, such as image generation and text generation in language modeling. A VAE~\citep{kingma2013auto,liu2017unsupervised} model includes two main parts: an encoder and decoder. The encoder first learns a conditional distribution of a latent variable $\mathbf{z}$, given observed data $\mathbf{x}$. The decoder then reconstructs data $\mathbf{x}$ from the generative distribution of latent code. However, due to intractable posterior inference, researchers often optimize the following evidence lower bound (ELBO).
\begin{equation}\small\label{eq:vae}
\mathcal{L}_{vae} =  \mathbb{E}_{q_\phi(\mathbf{z|x)}} [\log p_\theta(\mathbf{x|z})] - D_{KL} (q_\phi(\mathbf{z|x})||p(\mathbf{z}))
\end{equation}
where $p_\theta (\mathbf{x}|\mathbf{z})$ is a probabilistic \textit{decoder} parameterized by a neural network to reconstruct data $\mathbf{x}$ given the latent variable $\mathbf{z}$, and $q_{\phi} (\mathbf{z}|\mathbf{x})$ is a \textit{encoder} network that approximates the posterior distribution of latent variable $\mathbf{z}$ given data $\mathbf{x}$. In addition, $p(\mathbf{z})$ is a prior distribution of input data, such as Unit Gaussian. 

CVAEs~\citep{ivanov2018variational} build on the VAEs model by simply conditioning on $y$ (e.g., image generation given a label). CVAEs attempt to approximate the conditional distribution $p(x|y)$ given an input $y$. The variational lower bound of CVAEs can be expressed as
\begin{equation}\small \label{eq:cvae}
\begin{split}
\mathcal{L}_{cvae} = &  \mathbb{E}_{q_\phi(z|x,y)} \log p_\theta(x|z,y) \\
& - D_{KL} \big(q_\phi(z|x,y)||p_{\theta}(z|y) \big),
\end{split}
\end{equation}
where $p_{\theta}(z|y)$ is the prior probability distribution conditioned on $y$.

However, optimizing the VAEs and CVAEs model can lead to the KL-vanishing problem that KL-divergence becomes zero during model training in language modeling~\citep{bowman2015generating,liu2019cyclical}. Namely, the decoder only learns plain language without using the distribution of latent variable learned from the encoder. This is because the decoder is very powerful so that it can learn information from other paths, such as an attention mechanism~\citep{liu2019cyclical}.

\subsection{PID Control Algorithm}
\label{sec:pid_intro}
PID algorithm has been the most popular feedback control method in control theory, and it has been widely used in both physical systems~\cite{aastrom1993automatic} and computing systems~\cite{hellerstein2004feedback}. The basic idea of PID control is to calculate the error between a desired value (in this case, the desired KL-divergence) and the current actual value, then apply a correction in a direction that reduces that error. In its general form, the correction is the weighted sum of three terms; one is proportional to error (P), one is proportional to the integral of error (I), and one is proportional to the derivative of error (D). The general model of PID controller is defined by
\begin{equation}\label{eq:pid}
w(t) = K_p e(t) + K_i \int_0^t e(\tau)d\tau + K_d \frac{de(t)}{dt},
\end{equation}
where $w(t)$ is the output of the controller; $e(t)$ is the error between the actual value and the desired value at time $t$; $K_p, K_i$ and $K_d$ denote the coefficients for the P term, I term and D term, respectively.

\noindent
Since the derivative (D) term essentially computes the slope of the signal, when the signal is noisy it often responds more to variations induced by noise. Hence, we do not use D term based on established best practices in control of noisy systems. The resulting specialization of PID is called the PI controller.

In this paper, we propose a CVAEs model with the PI control algorithm to control the diversity and accuracy of generated sentences based on the input keywords and their order.


\section{Architecture of \textit{Apex}}
\label{sec:model}
We introduce the overall architecture of \textit{Apex} for controllable text generation. It combines the CVAEs model (as shown in Fig.~\ref{fig:apex}) with a PI controller (as shown in Fig.~\ref{fig:pid}) to manipulate the diversity and accuracy of generated text.

\subsection{Framework of CVAEs}
The CVAEs condition on the input keywords of items and their corresponding order, and then generate text to fulfill the keywords as specified in conditional encoder.
\subsubsection{Target Encoder and Conditional Encoder}
As shown in Fig.~\ref{fig:apex}, our target encoder is to encode the sequence of reference text via the Transformer model~\citep{vaswani2017attention}. Given a reference text $\mathbf{x}=(x_1,x_2, \dots, x_n)$ with $n$ elements, we embed each one in a low dimensional space as $\mathbf{E}=(e_1,e_2,\dots, e_n)$, where $e_i \in \mathbb{R}^f$ is the $i$-th column of an embedding matrix $\mathbf{D} \in \mathbb{R}^{f \times V}$, $f$ and $V$ denote the embedding size and the vocabulary size respectively. In addition, motivated by the BERT model~\citep{devlin2019bert}, we add a special token $[CLS]$ at the beginning of every reference text. The final hidden state
corresponding to this token is used as the aggregate sequence representation. Then the word embeddings of the reference text and the special token are fed into a $d$-dimensional Transformer, and it finally yields the output of hidden state $\mathbf{E}_{[CLS]}$, also denoted by $\mathbf{h}$.

For the conditional encoder, we also adopt the Transformer model to encode the input keywords $\mathbf{y}_a = (y_1,y_2, \dots, y_m)$ and their corresponding order $\mathbf{y}_p = (p_1,p_2,\dots,p_m)$ in the reference text. Similar to the target encoder, a special token $[CLS]$ is added at the beginning of the input keywords to aggregate sequence representation. In this encoder, the selected keywords of an item are first embedded in the low dimensional space, denoted by $\mathbf{U}=(u_1,u_2,\dots, u_m)$, where $u_j \in \mathbb{R}^f$ for $j=1,\dots,m$. Inspired by literature~\citep{gehring2017convolutional}, we embed their corresponding orders in the low dimensional space $\mathbf{O}=(o_1,o_2,\dots, o_m)$, where $o_j \in \mathbb{R}^f$. Then we combine their embeddings together with an element-wise addition, yielding element representations $\mathbf{C} = (u_1+o_1, u_2+o_2, \dots, u_m+o_m)$, where $\mathbf{C} \in \mathbb{R}^{f \times m}$. Finally, we feed both the embedding of the special token $[CLS]$ and the element representations $\mathbf{C}$ into the Transformer to obtain the hidden state $\mathbf{c}$.

After that, we concatenate the output hidden state of the above two encoders, and then feed it into the fully connected (FC) layers to generate the conditional distribution of latent variable, $\mathbf{z}$.


\subsubsection{Target Decoder}
The target decoder $G_\theta$ is to predict the target words via the Transformer given the input keywords and their order. Firstly, we concatenate the sampled data from the distribution of latent variable, $(\mu,\sigma)$, and the representation of each input keyword and the corresponding order, $[E_{1},E_{2},\dots E_{m}]$, from conditional encoder as the key ($K$) and value ($V$) of the Transformer. Namely, $K=V= [ z \oplus E_{1}, z \oplus E_{2},\dots z \oplus E_{m}]$. In addition, each word of reference sequence in the target decoder is embedded into a low dimensional space as the input query ($Q$). Finally, we can predict the target words based on $Q$, $K$ and $V$ in the Transformer decoder~\citep{vaswani2017attention}.


\subsection{Controllable KL-divergence using PI Control Algorithm}
We propose a linear PI control algorithm to manipulate the value of KL-divergence in Fig.\ref{fig:pid}, which has two advantages below:
\begin{itemize}[leftmargin=*]
\item Avert KL-vanishing problem. 
\item Achieve an adjustable trade-off between diversity and accuracy for generated text.
\end{itemize}

In language modeling, one challenge is that the VAEs and CVAEs model often suffer from the KL-vanishing problem due to a powerful decoder, as mentioned in Section~\ref{subsec:vae}. Our model also has the same problem because the Transformer decoder can predict the target words from the input keywords in the conditional encoder and its previous time steps. To mitigate KL-vanishing, some attempts~\citep{liu2019cyclical,bowman2015generating} have been made by researchers. One popular way is to add a hyperparameter, $w$, to the KL term in Eq.~\eqref{eq:cvae}, yielding:
\begin{equation}\label{eq:w_cvae}
\begin{split}
\mathcal{L}_{cvae} = & ~ \mathbb{E}_{q_\phi(z|x,y)} \log p_\theta(x|z,y) \\
& - w D_{KL} \big(q_\phi(z|x,y)||p_{\theta}(z|y) \big).
\end{split}
\end{equation}
\noindent
The basic idea is to set the weight $w$ to 0 at the beginning of model training, and then gradually increase it until to $1$ with an annealing method. The prior work, such as cost annealing and cyclical annealing~\citep{liu2019cyclical,bowman2015generating}, vary the weight in an open-loop fashion, without using feedback. When there exists a very powerful decoder such as Transformer, they cannot fully solve the KL-vanishing problem, as illustrated in Section~\ref{sec:evaluate}. To address this problem, one simple and direct method is to adjust the weight based on the actual output KL-divergence during model training. When the KL-divergence becomes too small, we need to reduce the weight to boost the KL-divergence. The PI control algorithm has such function, which can automatically tune the weight in the above VAE objective to stabilize the KL-divergence to a desired value (set point). 

In addition, KL-divergence plays an important role in the CVAEs because it affects the diversity and accuracy of generated text. When KL-divergence is large, the more diverse the generated text is, the lower its quality. In contrast, the generated text has higher reconstruction accuracy, but it is less diverse when KL-divergence is small. Thus, it is important to control KL-divergence to achieve a good trade-off between diversity and accuracy. This can be realized by PI control algorithm as well.

Fig.~\ref{fig:pid} illustrates the block diagram of our PI control algorithm. It samples the actual output KL-divergence during model training as the feedback to the PI controller. Then we compare the difference, $e(t)$, between the actual KL-divergence and desired one as the input of PI controller to calculate the weight $w$. According to Eq.~\eqref{eq:w_cvae}, when KL-divergence becomes smaller, the controller sets the weight $w$ to a smaller value in the objective function to reduce penalty for higher divergence; otherwise increases it until $1$, if KL-divergence becomes larger. During model training, we sample the KL-divergence at each training step in discrete time. Note that, as mentioned earlier, we do not use the full PID control, since the output of KL divergence is not very stable during model training with mini-batch data. Since the D term in PID control is a derivative term, it reacts to signal slope, which is often dominated by noise, and is therefore not productive. In order to improve the response time of control system, we adopt a linear PI control algorithm, as shown in Fig.~\ref{fig:pid}. The P term is proportional to the current error and I term sums (integrates) past errors with a sampling period $T_s=1$. Finally, the weight $w(t)$ is a sum of P term and I term below
\begin{equation}
w(t) = K_p e(t)+ K_i \sum_{i=0}^t e(i),
\end{equation}
where $0 \leq w(t) \leq 1$ and  $e(t)$ is the error between the desired KL-divergence and the output KL divergence at training step $t$; $K_p$ and $K_i$ are the \textit{negative} coefficients of the P term and I term, respectively.

Now, we briefly introduce how the designed PI control algorithm works. When error is large and positive (KL-diverge is below set point), both P term and I term becomes negative, so the weight $w(t)$ is limited to its lower bound $0$ that encourages KL-divergence to grow. When KL-divergence significantly exceeds the set point (called overshoot), the error becomes negative and large, both P term and I term become positive, so the weight becomes positive, causing KL-divergence to shrink. In both cases, the PI control algorithm dynamically tunes the weight to help KL-divergence approach the set point.

\begin{figure}[!tb]
\centering
\includegraphics[width= 3.1 in]{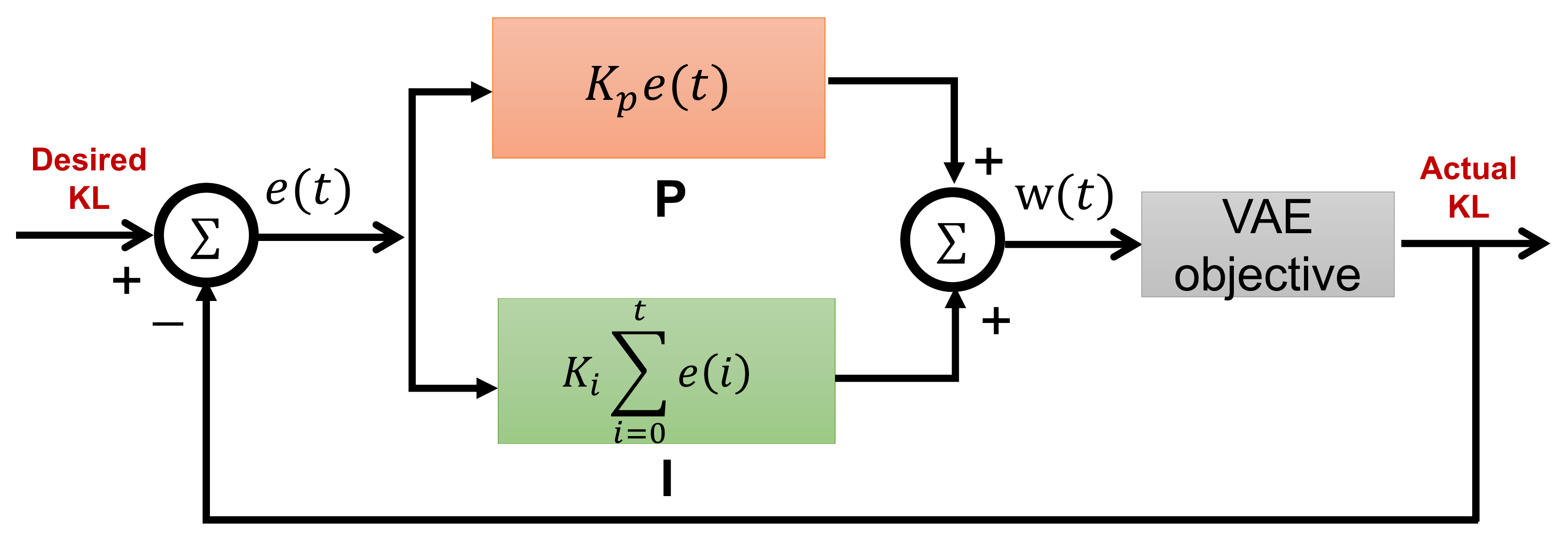}
\caption{PI controller component of the proposed \textit{Apex} for controllable KL-divergence.}
\vspace{-0.15 in}
\label{fig:pid}
\end{figure}

\subsection{Set Point Guidelines}
One question is: how to choose the desired value (set point) of KL-divergence for text generation?  For text generation, the weight $w$ in the VAE objective varies from $0$ to $1$, so we may choose the intermediate point, $w=0.5$, to run the CVAE models using cost annealing or cyclical annealing. The KL-divergence of CVAE model would converge to a specific value, $v_0$, after training many steps if not suffering from KL-vanishing (almost not). If it does suffer, we may choose a smaller weight $w$ until KL converges to a non-zero value, $v_0$. Then we can increase or decrease the desired KL-divergence to some extent based on the benchmark $v_0$. Note that, since our PI control algorithm is end-to-end dynamic learning method, users can customize the desired value of KL-divergence to meet their demand for different applications. For example, if users want to improve the diversity of generated text, they could choose a large set point of KL-divergence.

\subsection{Algorithm Summary}
We summarize the proposed PI controller in Algorithm~\ref{alg:pid}. Our PI algorithm updates the weight, $w(t)$, on the KL term in the objective using the sampled KL divergence at training step $t$ as feedback, as shown in Line $3$. Line $4$ computes the error $e(t)$ between sampled KL-divergence and set point. Line 5 to 12 is to calculate the weight $w$ based on error $e(t)$. Note that, Line 9 is a popular constraint in PID/PI design, called anti-windup~\citep{peng1996anti}. It effectively disables the integral term of the controller when controller output gets out of range, not to exacerbate the out-of-range deviation. 
\begin{algorithm}[!t]\small
\SetAlgoLined
\SetKwInOut{Input}{input}\SetKwInOut{Output}{output}
\Input{Desired KL divergence, $v_{KL}$, $K_p$, $K_i$, $\alpha$, training steps $T$}
\Output{weight on KL term $w(t)$ at training step $t$}
$I(0)=0$, $w(0)=0$\;
\For{$t=1$ \KwTo $T$}{
	sampled KL divergence $\hat{v}_{KL}(t)$ \;
	$e(t) \leftarrow v_{KL}-\hat{v}_{KL}(t)$ \;
	$P(t) \leftarrow K_p e(t)$ \;
	\uIf{ $0 \leq w(t-1) \leq 1$ }{
	$I(t) \leftarrow I(t-1) + K_i e(t)$
     }\Else{$I(t) \leftarrow I(t-1)$ // Anti-windup} 
     \tcp{calculate weight $w(t)$}
     $w(t) \leftarrow P(t)+I(t)$\;
     \If{$w(t)>1$}{
     $w(t) \leftarrow 1$} 
      \If{$w(t) < 0$}{
     $w(t) \leftarrow 0$}
     \Return  $w(t)$
}
 \caption{PI control algorithm.}\label{alg:pid}
\end{algorithm}

\section{Evaluation}
\label{sec:evaluate}
We evaluate the performance of \textit{Apex} on a real-world dataset. We first verify that the PI controller can totally avert KL vanishing. Then, we compare the performance of \textit{Apex} with baselines using automatic evaluation and human evaluation. We also show that our approach can generate text that obeys the order of keywords given by user input. Finally, we implement online A/B testing to demonstrate the good performance of the proposed \textit{Apex}.

\subsection{Datasets}
We collect a real world dataset of item descriptions and keywords from the Chinese E-commerce platform, Taobao. This dataset contains $617,181$ items and $927,670$ item text descriptions. Each item description is written by multiple human writers who help sellers write text to attract more customers. According to our statistics, each item on average has about $1.4$ descriptions written by humans and the average number of words in each description is $61.3$. In addition, there are about $14.75$ keywords per item on average. The total size of the vocabulary used is $88,156$. 

\subsection{Baselines}
We compare the performance of \textit{Apex} with the following baselines:
\begin{itemize}[noitemsep,topsep=0pt,leftmargin=*]
\item Apex-cost: This method uses cost annealing method instead of PI controller for \textit{Apex}.
\item Apex-cyclical: This method uses cyclical annealing method instead of PI controller for \textit{Apex}.
\item Seq2seq~\citep{duvsek2016sequence}: This is the seq2seq model with a maximum likelihood estimator~\citep{sutskever2014sequence}. This paper adopts the Transformer model instead of LSTM in the encoder and decoder for a fair comparison.
\item DP-GAN~\citep{xu2018diversity}: This conditional generator builds on the seqGAN, which uses the first sentence as conditional input.
\item ExpansionNet~\citep{ni2018personalized}: This model conditions on some aspects or keywords of items and user identity to generate personalized and diverse text.
\end{itemize}

\subsection{Experimental Settings}
We implement our experiments using Tensorflow through the Texar platform~\citep{hu2018texar}. We use a three-layer Transformer with default eight heads in the encoder and decoder. While transformer design often includes up to six layers, we find that three are sufficient for our purposes and use that number to improve efficiency. Besides, we adopt three-layer fully connected network with hidden size of $400$, $200$ and $100$ to learn the distribution of latent features. In our experiment, we use 80\%, 10\% and 10\% of the above dataset as the training, validation and testing data. We set the maximum number of words for each reference text to 100. The maximum number of input keywords is $50$. In addition, we set both the dimension of word embedding and order embedding to $512$. The learning rate is set to $0.0001$ with the exponential decay by a decay factor 0.9 every $1000$ training steps. Also, we set the batch size to $100$ for model training. Based on empirical PID tuning rules~\citep{silva2003stability}, we set the coefficient of PI algorithm, $K_p$ and $K_i$, to $-0.01$ and $-0.0001$, respectively. The sampling period, $T_s$, is set to 1 for our PI controller.


\begin{figure*}[!t]
  \centering
  \subfigure[KL divergence]{
  \includegraphics[width = 0.435\textwidth]{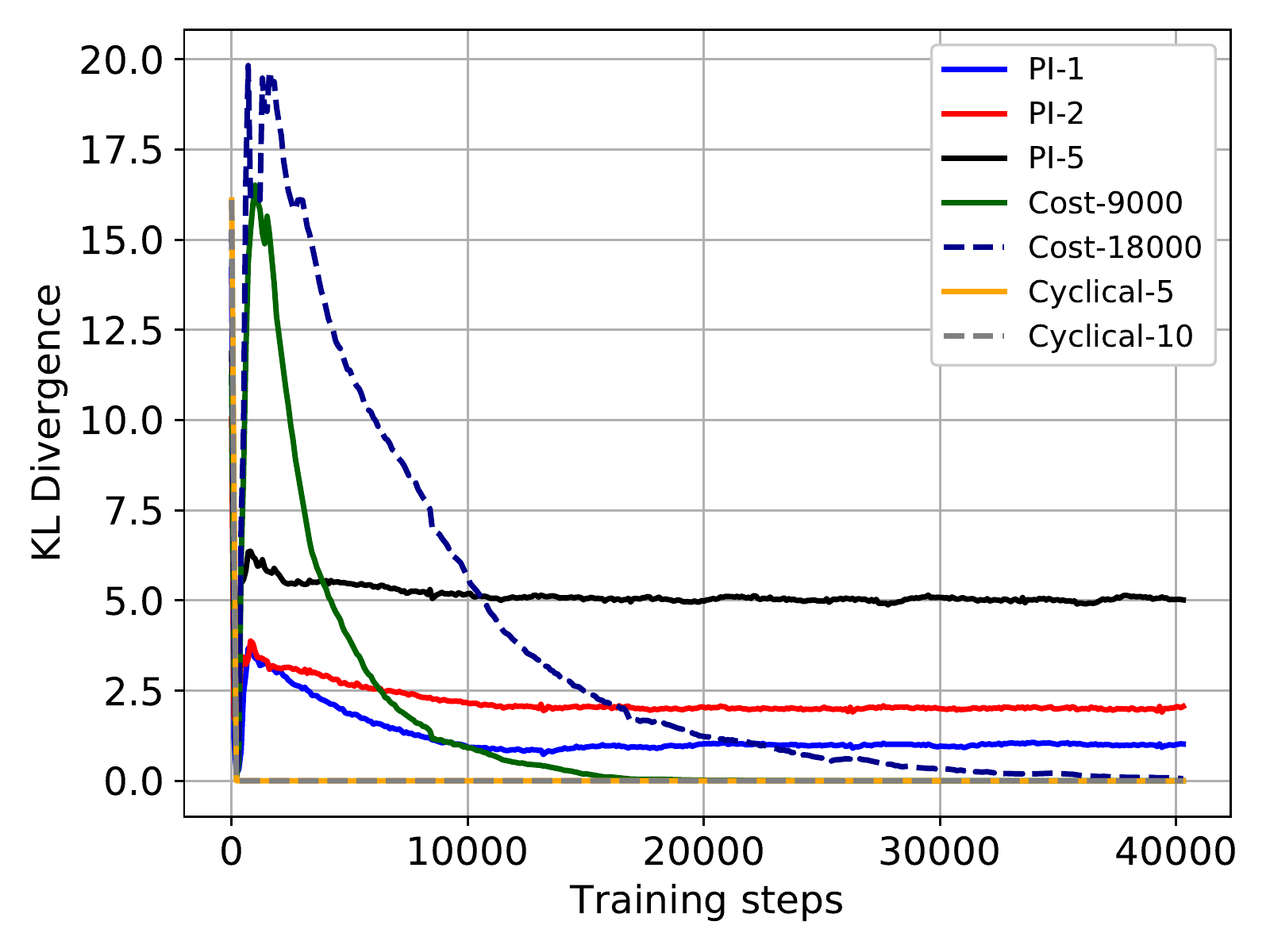}}
  \hskip -1ex
  \subfigure[Weight.]{
  \includegraphics[width =0.435\textwidth]{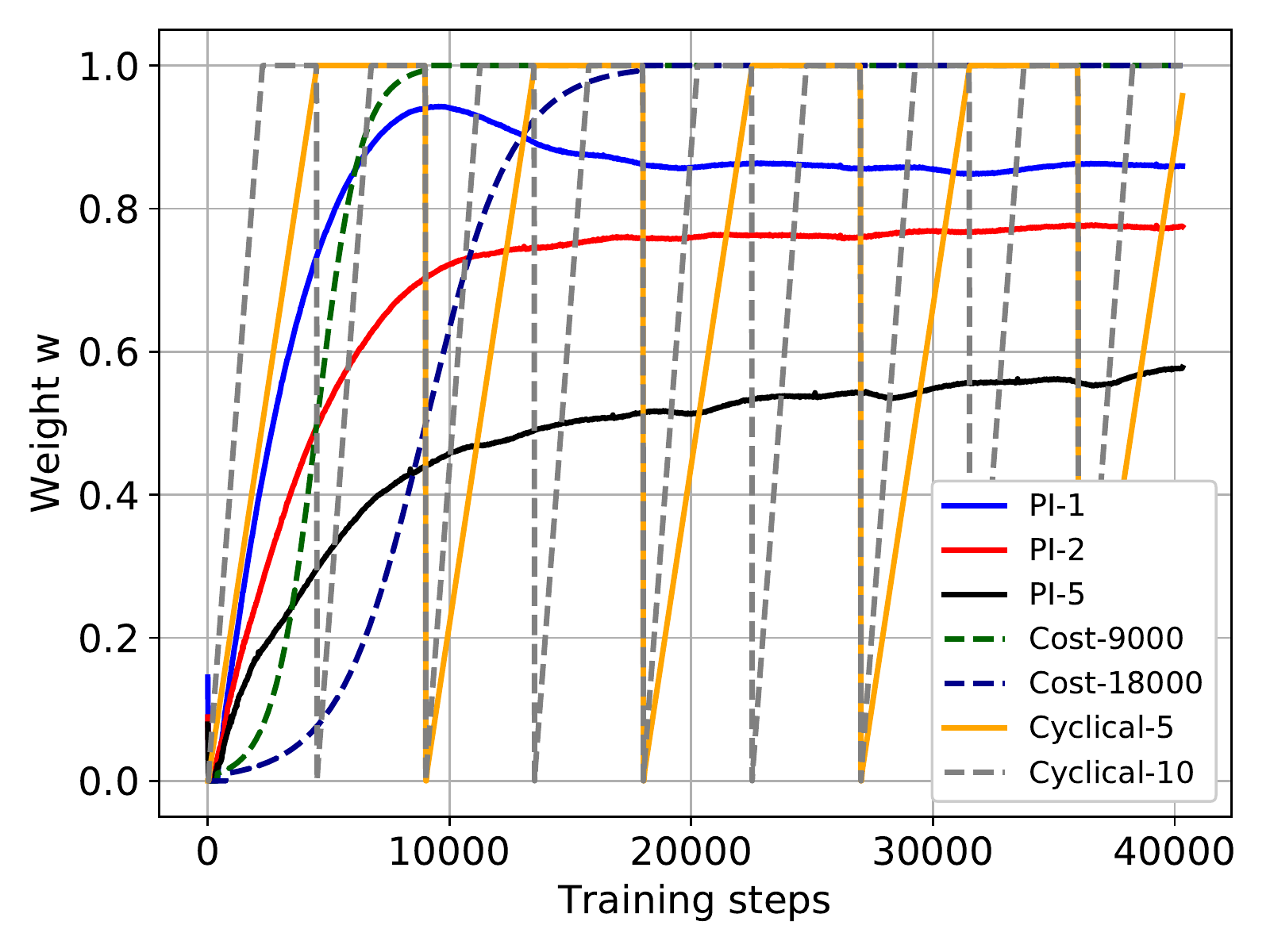}}
  \vskip -0.15in
  \caption{(a) illustrates comparison of KL divergence for different methods. We can observe that both cost annealing and cyclical annealing suffer from KL vanishing problem, while our \textit{Apex} totally averts KL vanishing. (b) Weight varies with the training step.}
  \label{fig:KL_vanish}
\end{figure*}

\begin{table*}[!thb]
\centering
\caption{Performance comparison for different methods using automatic metrics averaged over 5 random seeds. For Dis-$n$, ROUGE-L, METEOR: higher is better. Self-BLEU: lower is better. }
\begin{tabular}{ |l|c|c|c|c|c|c|c|c|}
 \hline
 Models & Dis-1 & Dis-2 &  Dis-3 & ROUGE-L & METEOR & self-BLEU-1 & self-BLEU-2 & self-BLEU-3 \\
 \hline
Apex-PI-5 &  \textbf{9.266K} & \textbf{127.51K} & \textbf{493.57K} & 0.2354 & \textbf{0.2593} & 0.9857 & 0.9471 & 0.8747 \\
Apex-PI-2 & 9.227K & 125.66K & 477.63K & 0.2358 & 0.2494 & \textbf{0.9841} & \textbf{0.9437} & \textbf{0.8674} \\
Apex-PI-1 & 9.108K & 125.39K & 476.40K & \textbf{0.2364} & 0.2538 & 0.9849 & 0.9454 & 0.8715\\
Apex-cost & 8.904K & 121.06K & 456.57K & 0.2344 & 0.2482  & 0.9853 & 0.9459  & 0.8719 \\
Apex-cyclical & 8.915K & 121.82K & 460.24K  & 0.2347 & 0.2486 & 0.9849 & 0.9444 & 0.8690 \\
Seq2seq & 8.648K & 112.31K & 427.22K & 0.2087  & 0.2399 & 0.9819 & 0.9480 & 0.8776 \\
ExpansionNet & 9.018K & 114.96K & 473.28K &  0.1951 & 0.1822 &0.9854& 0.9442 & 0.8700 \\
DP-GAN  & 8.124K & 102.18K & 413.78K & 0.1920 & 0.1949 & 0.9946 & 0.9849 & 0.9698 \\
\hline
\end{tabular}\label{tab:compare}
\end{table*}

\subsection{PI algorithm for KL vanishing}
We first conduct an experiment to demonstrate that the proposed PI algorithm (a variant of PID) can totally avert the KL vanishing. We compare it with the two representative approaches below:
\begin{itemize}[leftmargin=*]
\item Cost annealing~\citep{bowman2015generating}: This method firstly sets the weight of KL divergence term to 0 in the early stage, and then gradually increases it to $1$ using Sigmoid.
\item Cyclical annealing~\citep{liu2019cyclical}: This method splits the training process into $M$ (e.g., $5$) cycles and each increases the weight from 0 until to 1 using the linear function.
\end{itemize}

Fig.~\ref{fig:KL_vanish} illustrates that the KL divergence and its weight $w$ vary with the training steps for different methods. Note that, here PI-$v$ means we set the KL divergence to a desired value $v$ (e.g., 2) for our designed PI algorithm. We can observe from Fig.~\ref{fig:KL_vanish}(a) that when the training step is close to $100$, the KL divergence of CVAEs gradually decreases to $0$, which means the model suffers from KL vanishing. At that point, our PI algorithm automatically sets the weight to a small value to boost the KL divergence as shown in Fig.~\ref{fig:KL_vanish}(b). After $10,000$ training steps, the PI controller stabilizes the KL divergence at the customized set points, such as $1$, $2$ and $5$. Most importantly, we do not need to tune the PI parameters again for different set points of KL divergence in our experiment.

For the cost annealing method, we use two different hyper-parameters, $9000$ and $18000$, in the sigmoid function~\citep{bowman2015generating} to gradually increase the weight from $0$ until to $1$, as shown in Fig.~\ref{fig:KL_vanish}(b). From~\ref{fig:KL_vanish}(a), we can see that it does not suffer from KL vanishing problem at the beginning of training, because its weight is set to a small value in Fig.~\ref{fig:KL_vanish}(b). However, it gradually suffers from KL vanishing after increasing weight from 0 to 1.

For cyclical annealing, we also use different hyper-parameters (e.g., $5$ and $10$) about the number of cycles to do experiments. We discover that cyclical annealing suffers from KL vanishing a lot, because it blindly changes its weight without using the feedback of KL divergence. In Fig.~\ref{fig:KL_vanish}, it can be seen that when the KL divergence is decreasing at some points, its weight still increases, accelerating KL vanishing.

\subsection{Performance Comparison}
Next, we compare the performance of~\textit{Apex} with the baselines above, as shown in Table~\ref{tab:compare}. We adopt the commonly used metrics in NLP: Dis-$n$~\citep{xu2018diversity}, ROUGE-L~\cite{hovy2006automated}, METEOR~\cite{banerjee2005meteor}, and Self-BLEU~\cite{zhu2018texygen}, to evaluate their performance. Here \textit{Apex-PI-v} represents \textit{Apex} with PI algorithm to control the KL divergence to a desired value, such as $5$, $2$ and $1$. Besides, \textit{Apex-cost-anneal} and \textit{Apex-cyclical-anneal} represent our model but replace the PI algorithm with cost annealing and cyclical annealing to generate text, respectively. Table~\ref{tab:compare} illustrates that \textit{Apex} with PI controller has higher Dis-1, Dis-2 and Dis-3 but lower self-BLEU than the baselines. It means our \textit{Apex} can generate more diverse text than other methods. We also discover that the higher the KL divergence, the more diverse the generated text is. In addition, \textit{Apex} with PI algorithm has higher METEOR and ROUGE-L than the baselines, which means our method can generate accurate text based on keywords. Therefore, the proposed method can achieve a good trade-off between accuracy and diversity using PI control algorithm. We also show more examples in Appendix~\ref{app:example}.


\begin{table*}\footnotesize
\centering
\caption{Generated text by different models with different order of input keywords. Case 1 and case 2 show that \textit{Apex} can totally control the order of keywords. Case2-1 and 2-2 show that \textit{Apex} can generate diverse text with the same order of input keywords. However, the baselines cannot control the order of keywords.}
\begin{tabular}{|m{3.5cm}|m{12cm}|}
\hline
\multicolumn{2}{|l|}{
\begin{CJK*}{UTF8}{gbsn}\footnotesize
Item: 半身裙~(skirt)
\end{CJK*}
}  \\ \hline
\multicolumn{2}{|l|}{
\begin{CJK*}{UTF8}{gbsn}\footnotesize
Input keywords: 撞色, 流苏, 迷人, 裙摆 ~(Contrasting color, Tassel, Charming/Charm, Skirt's hemline)
\end{CJK*}
} \\ \hline

\multicolumn{2}{|c|}{
\begin{CJK*}{UTF8}{gbsn}\footnotesize
Model: Apex-PI-2
\end{CJK*}
}  \\ \hline
\begin{CJK*}{UTF8}{gbsn}\footnotesize
Case 1: \newline
撞色(1), 流苏(2), 迷人(3),裙摆(4) \newline Contrasting color(1), Tassel(2), \newline Charming(3), Skirt's hemline(4)
\end{CJK*} & \begin{CJK*}{UTF8}{gbsn}\footnotesize
这款半身裙,采用了{\color{red}撞色流苏}的设计,丰富了裙身的细节,增添了几分灵动的气息,散发出 {\color{red}迷人}的气息.而{\color{red}裙摆}的流苏设计,更是增添了几分俏皮感,更显别致. \newline This skirt is designed with {\color{red}contrasting-color tassel}, which enriches its details, adds a bit of instant style and shows a {\color{red}charming} fashion. Plus, the {\color{red}skirt's hemline} is decorated with tassel, which makes it more playful and chic.
\end{CJK*} \\ \hline
\begin{CJK*}{UTF8}{gbsn}\footnotesize
Case 2-1 \newline
裙摆(1), 流苏(2), 迷人(3), 撞色(4) \newline Skirt's hemline(1), Tassel(2), \newline Charming(3), Contrasting color(4)
\end{CJK*} & \begin{CJK*}{UTF8}{gbsn}\footnotesize
{\color{red}裙摆}处的{\color{red}流苏}装饰,为整体增添了几分灵动的美感, 行走间摇曳出灵动的{\color{red}迷人}姿态.而裙身的{\color{red}撞色}线条, 则是将活力的气息展现的淋漓尽致, 让人看起来更加的与众不同. \newline The {\color{red}tassel} decoration on the {\color{red}skirt's hemline} adds a little of instant style, which helps people show a glamorous and {\color{red}charming} gesture when walking. Also, the {\color{red}contrasting color} of this skirt shows the vitality, making it out of the ordinary.
\end{CJK*}
\\ \hline
\begin{CJK*}{UTF8}{gbsn}\footnotesize
Case 2-2 (same as above) \newline
裙摆(1), 流苏(2), 迷人(3), 撞色(4) \newline Skirt's hemline(1), Tassel(2), \newline Charm(3), Contrasting color(4)
\end{CJK*}
& \begin{CJK*}{UTF8}{gbsn}\footnotesize
这款半身裙的{\color{red}裙摆}采用{\color{red}流苏}的设计,展现出一种灵动飘逸的美感,行走间展现出女性{\color{red}迷人}的魅力. 裙身{\color{red}撞色}的设计,丰富了视觉效果,展现出女性十足的青春活力,彰显出女性的个性与柔美. \newline
The {\color{red}skirt's hemline} is designed with {\color{red}tassel}, showing an instant style and {\color{red}charm} of ladies when walking. Its {\color{red}contrasting color} enriches the visual effects, fully showing the vitality and youth of women as well as their personality and beauty.
\end{CJK*}         \\ \hline

\multicolumn{2}{|c|}{Model: Seq2seq} \\ \hline
\begin{CJK*}{UTF8}{gbsn}\footnotesize
撞色,迷人,裙摆,流苏\newline Tassel, Contrasting color, \newline Charming, Skirt's hemline
\end{CJK*}
& \begin{CJK*}{UTF8}{gbsn}\footnotesize
这款半身裙的亮点之处在于{\color{red}流苏}的设计,流苏的装饰,不仅展现出了与众不同的时尚品味,还增添了几分灵动的美感.而{\color{red}撞色}的格纹, 则展现出了一股复古的气息,彰显出了非凡的时尚品味,{\color{red}迷人}至极. \newline 
The fantastic part of this skirt is {\color{red}tassel}. With tassel as the decoration, the skirt not only shows distinctive fashion, but also provides a feeling of instant style. Moreover, its check pattern designed in {\color{red}contrasting color} shows vintage style and extraordinary fashion, which is very {\color{red}charming} and fantastic.
\end{CJK*}  \\ \hline
\multicolumn{2}{|c|}{Model: ExpansionNet} \\ \hline
 \begin{CJK*}{UTF8}{gbsn}\footnotesize
撞色,迷人,裙摆,流苏 \newline
Tassel, Contrasting color, \newline Charming, Skirt's hemline
\end{CJK*} 
& \begin{CJK*}{UTF8}{gbsn}\footnotesize
上半身的{\color{red}流苏}带设计,让裙身的艺术感更加上升,同时整体的裙身上又展示出秀气利落之感,同时大摆裙的{\color{red}裙摆}行走间随风摇曳飘逸,行走间凸显潮流气息. \newline
The design of {\color{red}tassel} can improve the artistic sense of the skirt. The entire skirt also shows a sense of elegance. Moreover, the {\color{red}skirt's hemline} sways with the wind when the wearer is walking, making it looks more fashionable.
\end{CJK*}   \\ \hline
\multicolumn{2}{|c|}{Model: DP-GAN} \\ \hline
\begin{CJK*}{UTF8}{gbsn}\footnotesize
撞色,迷人,裙摆,流苏 \newline
Tassel, Contrasting color, \newline Charming, Skirt's hemline
\end{CJK*}
& \begin{CJK*}{UTF8}{gbsn}\footnotesize
这质感的丝绒面料,让你在行走间散发出浪漫的气息.而裙身的{\color{red}撞色}设计,更是将女性的优雅气质展现的淋漓尽致.而裙身的{\color{red}流苏}装饰,则是以撞色的方式呈现出了一种独特的美感,让你在行走间散发出{\color{red}迷人}的气息. \newline 
This velvet fabric can make you send out a romantics taste. The design of {\color{red}contrasting color} on the skirt can totally show the elegance of ladies. In addition, its {\color{red}tassel} decoration shows a unique aesthetic feeling, making you look very {\color{red}charming} when walking.
\end{CJK*} \\ 
\hline
\end{tabular}\label{tab:control}
\end{table*}

\subsection{Accuracy on Controlling The Order of Keywords}
We then compare the accuracy in the control of the order of keywords for different methods. Our experimental results show the proposed \textit{Apex} achieves about 97\% accuracy for controlling the order of keywords in the generated sentences. The accuracy for seq2seq, DP-GAN and ExpansionNet, are 40\%, 16\% and 15\%, respectively. Thus, our method significantly outperforms the baselines in controlling the order of keywords. Table~\ref{tab:control} illustrates an example of product descriptions generated by~\textit{Apex} and baselines\footnote{The order of keywords may be different in English translation}. For case $1$ and case $2$, we can see that our \textit{Apex} can totally control the orders of keywords in the generated text. For case 2-1 and 2-2, we use the same order of keywords to generate text. It can be seen that \textit{Apex} can generate diverse results with the same input order of keywords. However, the baseline methods sometimes miss some keywords in the generated text. Hence, \textit{Apex} outperforms the baselines in terms of diversity and accuracy.

%
%

\subsection{Human Evaluation}
\label{sec:human}
We ask $25$ human graders to evaluate the relevance (accuracy) and fluency of generated sentences by different methods above. Specifically, we randomly sampled $200$ item text descriptions from each method identically, and mixed (shuffled) them together without telling graders which method generates which text. Then we provided the graders with the input keywords of items, ground truth and the corresponding generated results. The graders were asked to grade relevance and fluency separately on a three point scale based on the our instructions with demo examples. Score ``2'' means the generated text is relevant/fluent, and score ``1'' means it is somewhat relevant/fluent, while score ``0'' means it is not. To ensure the consistency of graders, we computed the Kendall’s tau coefficient to measure the correlation among the graders. The computed score is $0.82$, which indicates high consistency among graders. In addition, we use the two-tailed t-test to compare the scores of our proposed Apex-PI and the baselines. The evaluation results are statistically significant with $p < 0.05$.

\begin{table}[!hbt]\small
\centering
\caption{Scores of fluency and relevance by graders.}
\vskip -0.1in
\begin{tabular}{|l|c|c|c|}
 \hline
 Model & Fluency &Relevance  & Avg. \\
 \hline
Apex-PI-5 & 1.57 & 1.67 & 1.62 \\
Apex-PI-2 & \textbf{1.62} & 1.72 & \textbf{1.67}  \\
Apex-PI-1 & 1.60 & \textbf{1.72} & 1.66  \\
Apex-cost-anneal & 1.55 & 1.61 & 1.58 \\
Apex-cyclical-anneal  & 1.52 & 1.62  & 1.57 \\
Seq2seq & 1.55 & 1.57 & 1.56 \\
ExpansionNet  &1.56 & 1.50 & 1.53 \\
DP-GAN & 1.28 & 1.45 & 1.36 \\
 \hline
\end{tabular}\label{tab:human}
\vspace{-0.1in}
\end{table}


\begin{figure*}[!t]
\centering
\includegraphics[width=1\textwidth]{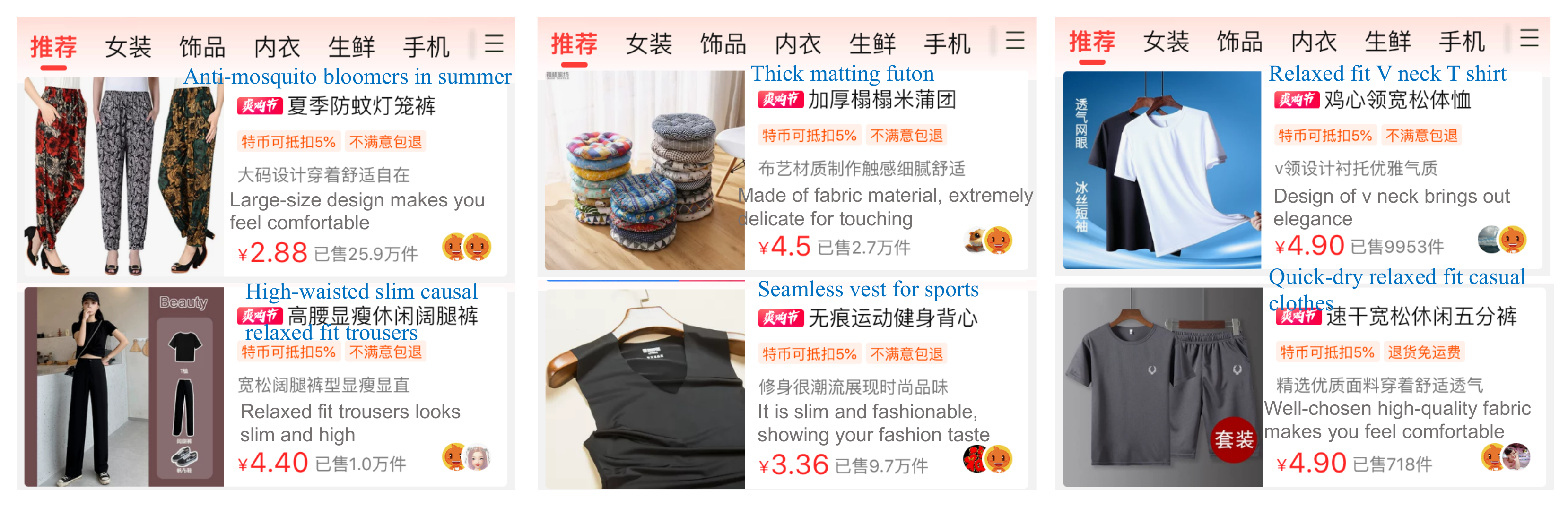}
\caption{Some examples of generated text for item recommendation in Taobao E-commerce platform. Production description is in blue color while recommendation reason is in darkgray color.}
\vspace{-0.1 in}
\label{fig:example}
\end{figure*}

Table~\ref{tab:human} illustrates the evaluation results by averaging the scores provided by graders. We can see that Apex-PI outperform the baselines in term of fluency and relevance. The fluency and relevance of Apex-PI-2 are better than those of Apex-PI-5. This is because a large KL divergence diversifies the generated text, but it lowers its quality. Besides, DP-GAN does not perform well because it is hard to stabilize the GAN. From this evaluation, we can conclude that KL divergence affects the accuracy of generated text. Thus, it is important to control the KL divergence to achieve the trade off between diversity and accuracy.

\subsection{Online A/B Testing}
We also do A/B test experiments to evaluate the performance of the proposed \textit{Apex}. We deploy \textit{Apex} in a real-world product in Taobao E-commerce platform for two applications: product description generation and item recommendation reason. For each item, we randomly choose one or two from some keywords with an average of 4 to generate different product descriptions and recommendation reasons. To get reliable and convincing evaluation results, the A/B test lasted for one week, from Aug.22th to Aug.28th, 2020. Fig.~\ref{fig:KL_vanish} illustrates some examples of generated text for item recommendation in Taobao. In the first A/B test experiment, we use \textit{Apex} to generate short production descriptions for recommended items, and then compare it with the long production descriptions by the existing method. The A/B test results show that \textit{Apex} can improve the Click-Though Rate (CTR) by 13.17\%. We also get the statistics about user demographics who are more likely to click, and the results show that our text generator improves CTR by 14.9\%, 10.9\% for older people and young people, respectively. For the second application, the proposed method generates one sentence to describe the advantages of the recommended items, and then compare it with user reviews and top-K items without reviews. The test result demonstrates that \textit{Apex} can achieve an improvement of about 6.89\% and 1.42\% in term of CTR over the baseline methods respectively. Therefore, the test experiments further verify the effectiveness of the proposed method.

\section{Related Work}
\label{sec:relatedwork}
In this section, we review the related work on text generation in recent years.

In early work, sequence to sequence (seq2seq)~\citep{sutskever2014sequence} models were widely used for text generation. However, seq2seq often overproduces high-frequency words and lacks diversity. In order to improve diversity, some follow-up work adopted deep generative models, such as VAE~\citep{hu2017toward,semeniuta2017hybrid} and generative adversarial networks (GANs)~\citep{goodfellow2014generative}. One of the most popular models is seqGAN~\citep{yu2017seqgan}, which leverages a discriminator and a generator to play a minimax game to generate text. Extensions were developed, such as DP-GAN~\citep{xu2018diversity},RevGAN~\citep{li2019towards}, RankGAN~\citep{lin2017adversarial} and LeakGAN~\citep{guo2018long}. However, these algorithms do not offer means to control the trade-off between diversity and accuracy.

Recent work attempted more controllable text generation. Some techniques control the generated text, conditioned on the attributes of users/items, key phrases, keywords and semantics of interest. For instance, Ni et al.~\citep{ni2018personalized} proposed ExpansionNet to generate personalized reviews by controlling user/item attributes and short phrases. In order to control story ending valence or keywords, researchers developed a recurrent neural network (RNN) based generation model~\citep{peng2018towards}. However, they did not consider controlling the order of keywords to diversity generated text.

There are also some works on production description generation in E-commerce~\cite{wang2017statistical,chan2019stick,gerani2014abstractive,lipton2015capturing}. Most of prior studies mainly leverage statistical methods with template to generate product descriptions. Recently, Chen et al.~\cite{chen2019towards,} developed a personalized product description
generation model based on knowledge graph. However, none of them is able to finely control the diversity or the order of keywords of generated text.

Different from existing work, we propose a novel controllable text generation model that can manipulate both the order of keywords to fit consumer interest, and the KL-divergence to achieve a controlled trade off between diversity and accuracy.


\section{Conclusion}
\label{sec:conlusion}
This paper developed a novel generative model, called \textit{Apex}, that combines CVAEs with a PI algorithm to controllably generate accurate and diverse text. Specifically, the CVAEs diversify text generation by controlling the order of input keywords. In addition, the designed PI algorithm manipulates the KL divergence to simultaneously achieve good accuracy and diversity, while averting the KL vanishing problem. The evaluation results on a real world dataset demonstrate that our method outperforms the existing generative models in terms of both diversity and relevance. Moreover, it can control the order of keywords in the generated sentences with about 97\% accuracy. Finally, we deploy \textit{Apex} in a real-world product in a Chinese E-commerce platform to do A/B testing. Results show that \textit{Apex} significantly improves CTR over the existing methods.

}

\begin{acks}
Research reported in this paper was sponsored in part by DARPA award W911NF-17-C-0099, DTRA award HDTRA118-1-0026, the Army Research Laboratory under Cooperative Agreement W911NF-17-20196, NSF CNS 18-15891, NSF CNS 19-32529, and Navy N00014-17-1-2783. The views and conclusions contained in this document are those of the author(s) and should not be interpreted as representing the official policies of the CCDC Army Research Laboratory, DARPA, DTRA, or the US government. The US government is authorized to reproduce and distribute reprints for government purposes notwithstanding any copyright notation hereon.
\end{acks}

\bibliographystyle{ACM-Reference-Format}
\bibliography{bibliography}
\onecolumn
\section{Examples of generated text by \textit{Apex}}
\label{app:example}
\begin{table}[!hb]\footnotesize
\centering
\caption{Generated text by \textit{Apex} based on the order of input keywords.}
\begin{tabular}{|m{3.5cm}|m{12.5cm}|}
\hline
Keywords in order & Generated Sentences \\
\hline
\begin{CJK*}{UTF8}{gbsn}\footnotesize
磨砂, 系带, 迷人 \newline scrubs, lace, charming
\end{CJK*} & \begin{CJK*}{UTF8}{gbsn}\footnotesize
这款皮衣外套采用了{\color{red}磨砂}质感的面料,手感柔软,穿着舒适;{\color{red}系带}的设计,勾勒出{\color{red}迷人}的身材曲线,更显女性的柔美气质,口袋的装饰,丰富了整体的造型,增添了实用性. \newline This leather jacket is made of {\color{red}scrubs} fabric, which feels soft and is comfortable to wear. The design of {\color{red}lace} outlines a {\color{red}charming} figure, more feminine temperament. The decoration of pocket enriches the overall shape and makes it more practical.
\end{CJK*} \\ \hline
\begin{CJK*}{UTF8}{gbsn}\footnotesize
毛衣, 高领, 保暖, 编织 \newline sweater, high-collar \newline warm, weaving
\end{CJK*} & \begin{CJK*}{UTF8}{gbsn}\footnotesize
这款{\color{red}毛衣}采用{\color{red}高领}设计,{\color{red}保暖}舒适,精致的{\color{red}编织}工艺,优雅大方,整体非常的时尚,简约百搭,尽显大气风范. \newline This {\color{red}sweater} is designed with a {\color{red}high collar}, {\color{red}warm} and comfortable, and its {\color{red}weaving} technology is elegant and fashionable, simple and versatile, which can show your grace.
\end{CJK*}
\\ \hline
\begin{CJK*}{UTF8}{gbsn}\footnotesize
红色, 七分袖, 搭配 \newline red, three quarter sleeve \newline matching.
\end{CJK*}
& \begin{CJK*}{UTF8}{gbsn}\footnotesize
{\color{red}红色}的连衣裙,尽显女性的知性与优雅.{\color{red}七分袖}的设计,适合各种不同的身形,{\color{red}搭配}一双高跟鞋,尽显女性的优雅. \newline
The {\color{red}red} dress shows women's grace and elegance. The {\color{red}three quarter sleeve} is suitable to different body shapes. It can show women's elegance when {\color{red}matching} it with a pair of high heels.
\end{CJK*} \\ \hline
\begin{CJK*}{UTF8}{gbsn}\footnotesize
连体, 条纹, 图案 \newline one-piece, stripe, pattern
\end{CJK*}
& \begin{CJK*}{UTF8}{gbsn}\footnotesize
这款{\color{red}连体}连衣裙,简约大方,经典{\color{red}条纹图案},时尚大方,时尚潮流,尽显女性的优雅气质,宽松的款式不挑身材,穿着舒适自在. \newline
This {\color{red}one-piece} dress is simple and generous, and the classic {\color{red}striped pattern} is stylish and fashionable, full of women's elegant temperament. The loose style does not pick your figure, and it is comfortable to wear.
\end{CJK*} \\ \hline
\begin{CJK*}{UTF8}{gbsn}\footnotesize
镂空, 破洞, 宽松 \newline hollow, hole, loose
\end{CJK*}
& \begin{CJK*}{UTF8}{gbsn}\footnotesize
{\color{red}镂空}的{\color{red}破洞}设计,让这款毛衣看起来更加的有个性,带有小小的性感,{\color{red}宽松}的款式,让你穿着舒适不受负重感,很好的凹造型. \newline
The {\color{red}hollow hole} design makes this sweater look more unique, with a little sexy, {\color{red}loose} style, making you comfortable to wear without a sense of weight, very good for posing.
\end{CJK*} \\ \hline
\begin{CJK*}{UTF8}{gbsn}\footnotesize
撞色, 宽松, 图案 \newline contrasting, loose, pattern
\end{CJK*}
& \begin{CJK*}{UTF8}{gbsn}\footnotesize
{\color{red}撞色}格纹的设计,经典大方,带有文艺气息,{\color{red}宽松}的版型,不挑身材,轻松驾驭;牛角扣的设计,丰富了整体的层次感,显得十分的复古,{\color{red}图案}的设计,显得精美别致. \newline
The design of the {\color{red}contrasting} checkered pattern is classic and elegant, with a good styling, {\color{red}loose} feeling, not picking the figure, easy to control, and the design of the horn buckle enriches the overall layering, giving a retro feeling, and the design of the {\color{red}pattern} is exquisite and chic.
\end{CJK*} \\ \hline
\begin{CJK*}{UTF8}{gbsn}\footnotesize
中长款, 立体, 纯色 \newline long style, stereoscopic, pure color
\end{CJK*}
& \begin{CJK*}{UTF8}{gbsn}\footnotesize
简约的{\color{red}中长款},{\color{red}立体}挺括,提升穿着品味.宽松的廓形,不挑身材,非常好驾驭.{\color{red}纯色}的设计,简约大方,营造出沉静秀气的少女感.简约的欧美风格,时尚百搭. \newline
Simple {\color{red}long style}, {\color{red}stereoscopic} and crisp, enhance the wearing taste. Loose silhouette, not picking the figure, is very easy to control. The {\color{red}pure color} design, simple and elegant, creating a calm and delicate girlish feeling, minimalist European style, fashionable and versatile.
\end{CJK*} \\ \hline
\begin{CJK*}{UTF8}{gbsn}\footnotesize
条纹, 下摆 \newline stripe, hem
\end{CJK*}
& \begin{CJK*}{UTF8}{gbsn}\footnotesize
经典的西装领设计,简约大方,将女性的干练与利落气质完美的展现出来.经典的{\color{red}条纹}拼接,个性独特,丰富了视觉效果,显瘦又显高挑.{\color{red}下摆}的A字设计,优化身材比例,展现出高挑的身材. \newline
The classic suit collar design, simple and elegant, perfectly showing women's competence and neat temperament. The classic {\color{red}stripe} stitching shows uniqueness, enriches the visual effect, and make you look thin and tall. The A-shaped design of the {\color{red}hem} optimizes the proportion of the figure and shows a tall figure.
\end{CJK*} \\ \hline
\begin{CJK*}{UTF8}{gbsn}\footnotesize
粉色, 圆领, 迷人 \newline pink, round neckline \newline charming
\end{CJK*}
& \begin{CJK*}{UTF8}{gbsn}\footnotesize
这款连衣裙采用了{\color{red}粉色}的色调,看起来清新甜美,带来了甜美的少女气息,经典的{\color{red}圆领}设计,修饰脸型,露出{\color{red}迷人}的锁骨,包臀的款式,凸显性感的女人味. \newline
This dress uses {\color{red}pink} tones, looks fresh and sweet, and brings a sweet girlish style. The classic {\color{red}round neckline} design modifies the face and reveals a {\color{red}charming} look clavicle, the sheath style highlighting sexy femininity.
\end{CJK*} \\ \hline
\begin{CJK*}{UTF8}{gbsn}\footnotesize
大牌,连帽 \newline big-name, hooded design
\end{CJK*}
& \begin{CJK*}{UTF8}{gbsn}\footnotesize
这款卫衣版型宽松,穿着舒适,且立体挺括,彰显{\color{red}大牌}风范. 胸前的logo刺绣,极具个性,也体现了品牌的魅力. {\color{red}连帽}的设计,休闲时尚,更能凸显休闲的气质. \newline
This hoodie is loose-fitting, comfortable to wear, and stereoscopic, showing its {\color{red}big-name} style. The logo embroidery on the chest is very personalized and also reflects the charm of this brand. {\color{red}Hooded design}, which is casual and fashionable, highlights the casual temperament.
\end{CJK*} \\ \hline
\begin{CJK*}{UTF8}{gbsn}\footnotesize
毛呢, 格纹, 翻领 \newline wool, plaid, lapel
\end{CJK*}
& \begin{CJK*}{UTF8}{gbsn}\footnotesize
{\color{red}毛呢}大衣,是冬季的不二之选.经典的{\color{red}格纹},给人一种复古的优雅感.{\color{red}翻领}设计,撞色线条点缀,给人一种优雅的感觉.衣身两侧的口袋装饰,方便实用. \newline
{\color{red}Wool} coat is the only choice in the winter. Classic {\color{red}plaid} gives a retro elegance. The {\color{red}lapel} design is embellished with contrasting color, giving an elegant feeling. Moreover, the pocket on both sides of the body is convenient and practical.
\end{CJK*} \\ \hline
\begin{CJK*}{UTF8}{gbsn}\footnotesize
优质, 系带, 超大 \newline high-quality, lace, large
\end{CJK*}
& \begin{CJK*}{UTF8}{gbsn}\footnotesize
这款风衣外套采用了{\color{red}优质}面料,手感柔软舒适,穿着体验非常好.领口采用{\color{red}系带}设计,增加了亮点,穿着时髦.{\color{red}超大}口袋,实用方便. \newline
This windbreaker jacket is made of {\color{red}high-quality} fabrics and feels soft and comfortable, giving pleasant wearing experience. The design of {\color{red}lace} in neckline is fashionable. Plus, its {\color{red}large} pocket is practical and convenient.
\end{CJK*} \\ \hline
\end{tabular}\label{tab:app-gen}
\end{table}


\end{document}